\documentclass[journal,twoside,web]{ieeecolor}
\usepackage{generic}
\usepackage{cite} 
\usepackage{multirow}
\usepackage{subcaption} 
\usepackage{amsmath,amssymb,amsfonts}
\usepackage{algorithmic}
\usepackage{graphicx}
\usepackage{algorithm,algorithmic}
\usepackage{textcomp}
\usepackage{tabularx, booktabs}
\usepackage{bbm}

\usepackage[hypertexnames=false]{hyperref}
\hypersetup{
    colorlinks=true,
    allcolors=blue,
    citecolor=blue
}
            
\def\BibTeX{{\rm B\kern-.05em{\sc i\kern-.025em b}\kern-.08em
    T\kern-.1667em\lower.7ex\hbox{E}\kern-.125emX}}
\begin{document}
\title{AI-Driven Prediction of Cancer Pain Episodes: A Hybrid Decision Support Approach}
\author{Yipeng Zhuang, \IEEEmembership{Member, IEEE}, Yifeng Guo, \IEEEmembership{Member, IEEE},  Yuewen Li, Yuheng Wu, Philip Leung‐Ho Yu, Tingting Song, Zhiyong Wang, Kunzhong Zhou, Weifang Wang, and Li Zhuang
\thanks{We thank the associate editor and the reviewers for providing valuable and constructive comments which greatly improved our paper. This work was supported by the National Natural Science Foundation of China (82260589 and 82460540), the Key Project of the Basic Joint Special Program of Yunnan Provincial Science and Technology Department - Kunming Medical University (202401AY070001-014), the Construction of First-Class Discipline Team of Kunming Medical University (2024XKTDPY08), and the Medical and Health Talent Special Project of the "Support Program for Developing Talents in Yunnan" (Grant No. CZ0096-901895). (Y. Zhuang, Y. Guo, and Y. Li contributed equally to this work as co-first authors.) (Corresponding author: L. Zhuang.)}
\thanks{Y. Zhuang is with the Department of
Orthopaedics and Traumatology, The
University of Hong Kong, Hong Kong (email: yipengzh@hku.hk).}
\thanks{Y. Guo, and P.L.H. Yu are with the Department of Statistics and Actuarial Science, Faculty of Science, The University of Hong Kong, Hong Kong (email: gyf9712@connect.hku.hk; plhyu@eduhk.hk).}
\thanks{Y. Li, T. Song, Z. Wang, K. Zhou, W. Wang, and L. Zhuang are with the Rehabilitation and Palliative Medicine Department, Peking University Cancer Hospital Yunnan Hospital, The Third Affiliated Hospital of Kunming Medical University, Kunming, China (email: 20211312@kmmu.edu.cn; medicalsong@163.com; 20201184@kmmu.edu.cn;  20231409@kmmu.edu.cn; 20231408@kmmu.edu.cn; zhuangli@kmmu.edu.cn).}
\thanks{Y. Wu is with the Department of Biomedical Engineering, City University of Hong Kong, Hong Kong (email: yuhengwu7-c@my.cityu.edu.hk).}}
\maketitle

\begin{abstract}
Lung cancer patients frequently experience breakthrough pain episodes, with up to 91\% requiring timely intervention. To enable proactive pain management, we propose a hybrid machine learning and large language model pipeline that predicts pain episodes within 48 and 72 hours of hospitalization using both structured and unstructured electronic health record data. A retrospective cohort of 266 inpatients was analyzed, with features including demographics, tumor stage, vital signs, and WHO-tiered analgesic use. The machine learning module captured temporal medication trends, while the large language model interpreted ambiguous dosing records and free-text clinical notes. Integrating these modalities improved sensitivity and interpretability. Our framework achieved an accuracy of 0.876 (48h) and 0.917 (72h), with improvements in sensitivity of 10.6\% and 10.7\%, respectively, attributable to large language model augmentation. This hybrid approach offers a clinically interpretable and scalable tool for early pain episode forecasting, with potential to enhance treatment precision and optimize resource allocation in oncology care.

\end{abstract}


\section{Introduction}
\label{sec:introduction}
Lung cancer remains the leading cause of cancer-related mortality globally, with pain affecting 42.2\% of patients and 91\% reporting moderate-severe intensity~\cite{nurwidya_pain_2016}. The World Health Organization (WHO) recommends a stepwise analgesic ladder for cancer pain management, progressing from non-opioid analgesics (e.g., NSAIDs) to weak (e.g., codeine) and strong opioids (e.g., morphine)~\cite{world1996cancer}. However, despite these guidelines, effective pain control remains challenging due to limitations in provider training and restricted access to opioids in many settings~\cite{simmons_clinical_2012}.

Predicting pain episodes in hospitalized patients holds significant clinical value by enabling timely analgesic adjustments, preventing breakthrough pain, and optimizing resource utilization. This is especially critical in inpatient oncology, where lung cancer patients experience among the highest rates of moderate to severe pain. Delays in pain control can lead to escalated opioid use, reduced functional status, and diminished quality of life. In clinical practice, a 72-hour observation window is often used to evaluate the effectiveness of pain management strategies, consistent with WHO and institutional guidelines~\cite{md_anderson_cancer_center_cancer_2025, talmi_pain_1997, mercadante_opioid_1999}.

Machine learning (ML) methods have been used to forecast cancer pain and other symptoms using structured EHR data. For example, in musculoskeletal pain, baseline disability and early symptom changes can predict healthcare utilization~\cite{lentz_prediction_2018}; in migraine, ML has been used to classify pain trajectories based on self-reported features~\cite{galvez-goicurla_cluster-then-classify_2022}; and in end-of-life care, structured electronic health record (EHR) models have forecasted pain at hospitalization and shift levels~\cite{lodhi_predictive_2015}. ML models, including logistic regression, random forests, and neural networks, have also been applied to predict cancer-related symptoms such as pain, fatigue, and depression~\cite{alfayez_predicting_2021, zeinali_machine_2024}. However, most existing approaches treat symptoms and medication as static variables, ignoring the temporal dynamics essential for acute pain prediction.

While recent time-series models have demonstrated the ability to capture temporal dynamics in structured data, such as numerical rating scale (NRS) scores for pain~\cite{bang_clinical_2023}, these approaches still face critical limitations. Most of them rely exclusively on structured inputs, overlooking valuable unstructured information like free-text medication logs that often contain key contextual cues such as irregular opioid dosing or rescue medication use. Additionally, temporally-aware models often treat medications and interventions as static covariates or aggregate summaries, failing to reflect real-time adjustments in patient care. For example, previous work~\cite{chae_prediction_2024} incorporated longitudinal nursing notes to monitor symptom progression, but did not target pain prediction or leverage multi-modal data integration. These gaps reduce the clinical applicability of current models, especially in cases where documentation is inconsistent or ambiguous.

Large language models (LLMs) have demonstrated the capacity to extract structured information from unstructured clinical narratives, including medication dosing patterns, rescue analgesic use and temporal relationships. These models are increasingly integrated into hospital workflows worldwide, where they support medication reconciliation, symptom tracking and clinical documentation across diverse health systems \cite{thirunavukarasu2023large}. However, when used independently, LLMs often yield poorly calibrated outputs on mixed-format clinical datasets. This limitation is particularly pronounced for structured tabular inputs, where numerical values and relational dependencies may be misinterpreted, leading to unreliable predictions~\cite{sui2024table, yan2025small}.

The role of multimodal data fusion in biomedical applications has been systematically characterized by Stahlschmidt et al. \cite{stahlschmidt2022multimodal}, who classified fusion strategies into early, intermediate, and late paradigms. Unlike conventional late-fusion approaches that unconditionally aggregate all modality outputs, our framework introduces a decision-level fusion mechanism: the LLM is invoked only when the ML classifier's output falls within a predefined marginal-confidence interval, thereby selectively combining modalities rather than merging them indiscriminately. This design reduces computational overhead and limits LLM-induced noise in high-confidence cases. In a related direction, Swaminathan et al. \cite{swaminathan2024selective} showed that selective prediction for unstructured clinical data can improve reliability by restricting automated decisions to cases with high confidence, a principle that closely parallels our uncertainty-aware routing mechanism.

To address these limitations, we introduce a hybrid pipeline integrating:
\begin{itemize}
    \item \textbf{Dynamic pain episode prediction}: Pain episodes were forecasted by integrating static EHR data with temporal features extracted from clinical records.
    
    \item \textbf{LLM augmentation}: DeepSeek-R1~\cite{guo2025deepseek} was employed for contextual interpretation of clinical notes, enhancing prediction accuracy in marginal cases.

    \item \textbf{Clinically actionable horizons}: The 48- and 72-hour predictions, aligned with WHO-recommended reassessment intervals~\cite{world1996cancer}, achieved high accuracy and sensitivity, supporting their potential clinical utility.
\end{itemize}

Our retrospective study included 304 lung cancer inpatients. The dataset includes both structured EHR medication records and unstructured nursing documentation, offering a unique opportunity to evaluate advanced AI-driven and time-aware pain forecasting methods in real-world clinical settings to support proactive and context-aware decision making.

\section{Methods}
\subsection{Study Design and Setting}
This retrospective study was conducted at the third Affiliated Hospital of Kunming Medical University, where data from 304 lung cancer inpatients were initially collected between January 2020 and December 2023. The study protocol was approved by the institutional review board (KYLX2023-193), and informed consent was waived due to the retrospective nature of the analysis. Participants must have pathologically confirmed lung cancer along with cancer-related pain. They should have undergone at least three pain assessments, with a Numeric Rating Scale (NRS) score of 1 or higher, or an equivalent pain assessment method. Additionally, records of analgesic medication use must be available. Participants must not have received radiotherapy within the past 15 days. They should also be free from recent infections or fever. Individuals were excluded if they had severe heart, liver, or kidney dysfunction, including liver function classified as Child-Pugh grade III, or suffered from nephrotic syndrome. Patients with concurrent physical pain conditions, loss of consciousness impairing pain assessment, or allergies to morphine-based medications were also excluded. Patients were excluded if they had over 30\% missing clinical data. After applying exclusion criteria, 266 patients were included in the final analysis.

\subsection{Data Preprocessing}
\label{sec:data_preprocess}
We systematically analyzed the natural language text in the dataset that recorded the intensity and duration of pain. Using a rule-based extraction algorithm, qualitative pain descriptions were converted into numerical pain scores corresponding to the 24-, 48-, and 72-hour time intervals after admission. A threshold of 4 on the numerical rating scale (NRS) was used for binary classification, with scores $\geq 4$ classified as \textit{pain positive} and scores $< 4$ as \textit{pain negative}~\cite{bang_clinical_2023}.

The framework performs conditional risk prediction under the assumption of maintained current medication. To prevent information leakage, strict temporal cutoffs were enforced. For the 48-hour prediction task, only medication records and clinical observations documented within the first 24 hours of admission were used as predictor variables; the outcome was assessed independently at the 48-hour mark. Similarly, for the 72-hour prediction, predictor features included medication and clinical data from the first 24 and 48 hours, while the outcome was measured at 72 hours. No information from the prediction target window (i.e., beyond the feature extraction cutoff) was included in the feature set. Clinical notes and chief complaints were also filtered by their documented timestamps to ensure adherence to this temporal boundary. 

Categorical variables representing pathological subtypes were encoded into ordered categories according to clinical relevance to avoid sparsity issues. 

Specifically, adenocarcinoma, squamous cell carcinoma, neuroendocrine carcinoma, and soft tissue sarcoma were encoded as distinct ordinal variables. Population demographic covariates such as smoking history and gender were one-hot encoded to mitigate potential biases from misclassification. 

Additionally, drug data were extracted from the medication records. Based on the WHO analgesic ladder, drugs were classified into three levels: potent opioids, moderate opioids, and non-opioid analgesics, as shown in Table~\ref{tab:drug_classification}\cite{world1996cancer, caraceni2012use}. The usage of each level within 24 hours and 48 hours were binarized to generate nine new binary features. Missing drug data were imputed using backward filling, and dosage values were log-transformed to normalize distribution.

To evaluate the generalizability of the hybrid model, an independent temporal validation cohort (n = 130) was assembled from patients enrolled between Aug 2025 and Feb 2026, a period entirely non-overlapping with the training and test sets.

\begin{table}[ht]
\centering
\caption{Drug Classification}
\label{tab:drug_classification}
\begin{tabularx}{\columnwidth}{>{\centering\arraybackslash}p{0.1\linewidth} >{\raggedright\arraybackslash}p{0.35\linewidth} >{\raggedright\arraybackslash}p{0.45\linewidth}}
\toprule
\textbf{Tier} & \textbf{Ladder} & \textbf{Drugs} \\
\midrule
3 & Strong opioids & Morphine, fentanyl, oxycodone \\
2 & Moderate opioids & Codeine, tramadol \\
1 & Non-opioid analgesics & Ibuprofen, acetaminophen \\
\bottomrule
\end{tabularx}
\end{table}

\subsection{Patient Characteristics}
{ 
Demographic and clinical characteristics were compared between the pain-negative and pain-positive groups. Representative variables with low missingness and clinical relevance are summarized in Table~\ref{tab:demographics}. Several laboratory markers differed significantly between groups, including higher AST in the pain-positive group (25.00 vs 21.00), lower albumin (38.95 vs 42.00), lower total protein (67.70 vs 71.65), and higher lactate dehydrogenase (235.00 vs 206.30). In addition, the pain-positive group showed lower potassium (4.05 vs 4.23) and lower sodium levels (139.00 vs 140.00). Because several laboratory variables exhibited skewed distributions, continuous variables are summarized as median with first and third quartiles, and between-group comparisons were performed using the Wilcoxon rank-sum test.

\begin{table}[ht]
\centering
\caption{Demographics and Clinical Characteristics by NRS Pain Score Group}
\label{tab:demographics}
\footnotesize
\renewcommand{\arraystretch}{1.2}
\setlength{\tabcolsep}{3pt}
\begin{tabularx}{\linewidth}{|l|>{\raggedright\arraybackslash}X|>{\raggedright\arraybackslash}X|c|c|}
\hline
\textbf{Characteristic} & \textbf{NRS$<4$} (n=125) & \textbf{NRS$\geq$4} (n=141) & \shortstack{\textbf{Missing} \\ \textbf{Rate}} & \textit{p}-value \\
\hline
\multicolumn{5}{|l|}{\textbf{Demographics}} \\
Age (y) & 57.00 \newline (52.00, 66.00) & 58.00 \newline (51.00, 64.25) & 2.3\% & 0.6734 \\
Sex* & & & 5.6\% & \textbf{0.0164} \\
\quad Male & 49 (42.2\%) & 37 (27.4\%) & & \\
\quad Female & 67 (57.8\%) & 98 (72.6\%) & & \\
Smoking History* & & & 11.7\% & 0.2747 \\
\quad Yes & 44 (38.6\%) & 38 (31.4\%) & & \\
\quad No & 70 (61.4\%) & 83 (68.6\%) & & \\
\hline
\multicolumn{5}{|l|}{\textbf{Clinical characteristics}} \\
\multicolumn{5}{|l|}{Liver Function Tests} \\
AST & 21.00 \newline (17.95, 27.00) & 25.00 \newline (19.00, 33.00) & 1.9\% & \textbf{0.0121} \\
\multicolumn{5}{|l|}{Hematology} \\
MCH & 30.40 \newline (29.40, 32.17) & 30.00 \newline (28.50, 31.60) & 1.9\% & \textbf{0.0229} \\
\multicolumn{5}{|l|}{Biochemistry} \\
Albumin & 42.00 \newline (37.80, 45.00) & 38.95 \newline (33.72, 42.35) & 2.3\% & \textbf{0.0001} \\
Total Protein & 71.65 \newline (66.00, 76.00) & 67.70 \newline (60.00, 73.00) & 1.9\% & \textbf{0.0002} \\
LDH & 206.30 \newline (172.00, 261.75) & 235.00 \newline (195.00, 321.40) & 1.9\% & \textbf{0.0027} \\
Potassium & 4.23 \newline (3.93, 4.50) & 4.05 \newline (3.87, 4.38) & 1.9\% & \textbf{0.0204} \\
Sodium & 140.00 \newline (137.75, 142.00) & 139.00 \newline (136.00, 141.00) & 1.9\% & \textbf{0.0469} \\
\hline
\end{tabularx}
\vspace{1ex}
\begin{minipage}{0.95\linewidth}
\footnotesize
\textbf{Note.} Values are presented as median (Q1, Q3) or n (\%), as appropriate. Percentages for categorical variables were calculated among non-missing observations within each group. Continuous variables were compared using the Wilcoxon rank-sum test, and categorical variables were compared using the Fisher's exact test. Variable-specific missingness is reported separately. AST = aspartate aminotransferase; MCH = mean corpuscular hemoglobin; LDH = lactate dehydrogenase.
\end{minipage}
\end{table}

\subsection{Model Development}
To develop a clinically actionable pain prediction model, we implemented a hybrid framework combining structured machine learning techniques with large language model capabilities, as shown in Figure~\ref{fig:pipeline}. This approach was designed to capture both temporal dynamics in analgesic usage and the complexity of unstructured clinical documentation.

\begin{figure*}[t] 
    \centering
    \includegraphics[width=\textwidth]{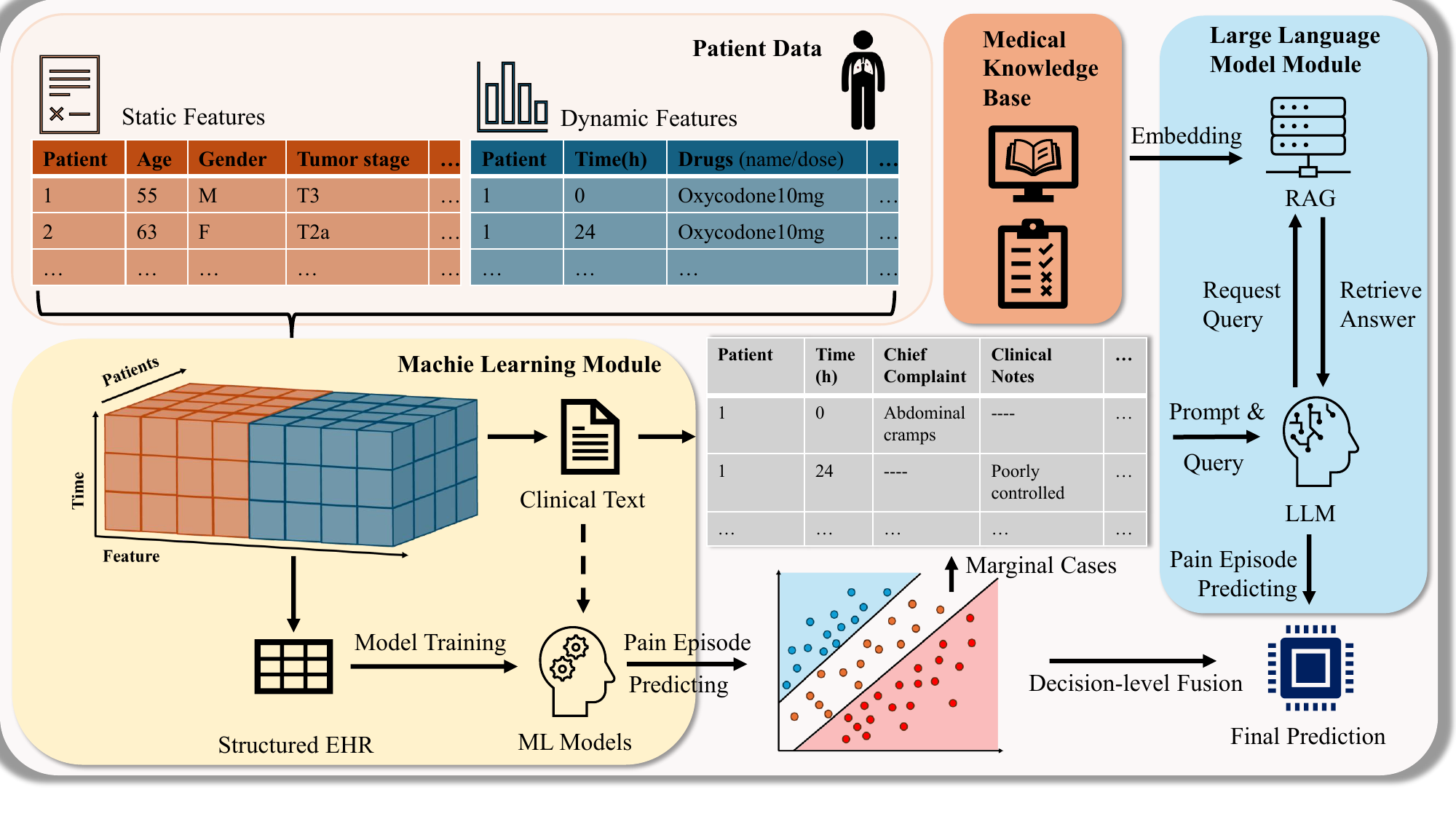}
    \caption{Overview of the hybrid ML–LLM prediction pipeline. Structured EHR features are processed by the ML classifier; cases with marginal confidence ($\alpha<p_{ML}<\beta$) are routed to the RAG-augmented LLM for secondary assessment. Final predictions are produced via decision-level fusion.}
    \label{fig:pipeline}
\end{figure*}

\subsubsection{Machine Learning Module}
We first developed a suite of supervised learning models to predict pain occurrence at 48-hour and 72-hour horizons. Algorithms evaluated included Random Forest (RF), Logistic Regression (LR), Support Vector Machines (SVM), Extreme Gradient Boosting (XGBoost), CatBoost, Extra Trees, Lasso, Gradient Boosting, LightGBM, and a Stacking Ensemble (combining RF, LR, GB and XGBoost as meta-learners). All models were trained using stratified 5-fold cross-validation to ensure balanced representation of pain outcomes and enhance generalizability.

As the prediction window extended from 48 to 72 hours, the number of pain-positive cases progressively declined, leading to an increasingly imbalanced class distribution. This imbalance not only posed challenges for model training but also increased the risk of false positives in later stages which may lead to unnecessary interventions that cause overtreatment or patient discomfort. To address this, we applied the synthetic minority oversampling technique (SMOTE) when the minority-to-majority class ratio fell below 0.3. This approach enhanced the model’s ability to learn from scarce positive cases while maintaining control over false positive rates in clinically sensitive timeframes. Feature importance was extracted for models supporting it (e.g., tree-based models, Lasso), revealing key predictors such as opioid dosage timing, cumulative analgesic exposure, and recent pain-related complaints.

In the second phase, we incorporated dynamic medication dosage variables extracted from unstructured medication records. Due to the large variety of drug names, we followed clinical medication guidelines to classify analgesics into three tiers: strong opioids, moderate opioids, and non-opioid analgesics, as we mentioned in Table~\ref{tab:drug_classification}. Based on this, we constructed temporal medication dosing variables for each patient, summarizing their analgesic exposure within the first 24 hours and 48 hours of admission. These variables were then added to the models to predict pain occurrence at 48h and 72h respectively.

\subsubsection{Large Language Model Module}

To augment our structured machine learning framework, we integrated DeepSeek-R1-Distill-Qwen-14B (14B parameters; $max-tokens=2048$, $temperature=0.4$)~\cite{guo2025deepseek}. We selected this distilled variant to optimize the trade-off between advanced reasoning capabilities and computational overhead. This selection was driven by three operational imperatives: (1) its inherent chain-of-thought reasoning facilitates the context-aware interpretation of irregular medication logs and longitudinal clinical narratives; (2) its open-weight architecture permits secure, on-premise deployment, strictly satisfying institutional data privacy mandates; and (3) its foundational pre-training on Qwen 2.5 ensures native proficiency with Chinese clinical corpora.To ensure high-fidelity LLM inputs, unstructured clinical texts underwent a rigorous preprocessing pipeline. 
We employed regular expressions to strip non-clinical metadata and mapped domain-specific abbreviations to full clinical entities using a custom lexicon. 
Notes containing fewer than 10 characters or consisting entirely of duplicated template text were filtered out as non-informative. Medication records were normalized by mapping drug brand names to their generic equivalents using a hospital-specific formulary mapping table. These preprocessing steps were applied consistently across all patient records prior to prompt construction.
Similarly, chief complaints may reflect early signs of breakthrough pain or patient discomfort, while clinical notes provide nuanced, temporally rich descriptions of symptom progression and therapeutic decisions. 

We implemented a Retrieval-Augmented Generation (RAG) framework to enhance the contextual reasoning capabilities of the base LLM. The RAG architecture consists of two primary components: a retriever and a generator. The retriever is responsible for identifying relevant external documents from a domain-specific knowledge base $D$, while the generator synthesizes the retrieved content with the input prompt to produce a final response. Formally, given an input query $g$, the retriever selects a set of top-$k$ documents $\{d_1, d_2, \dots, d_k\} \in D$ using a dense vector similarity search based on cosine similarity:
\begin{align}
\text{sim}(q,d_i) = \frac{\langle \phi(q), \phi(d_i) \rangle}{\|\phi(q)\| \cdot \|\phi(d_i)\|},
\end{align}
where $i = 1,2,\dots,k$, and $\phi(\cdot)$ denotes a BERT-based embedding function that maps queries and documents into a shared semantic vector space. 
{ 
Specifically, the knowledge base D comprised three categories of documents: (1) WHO cancer pain management guidelines \cite{world1996cancer}, (2) NCCN Adult Cancer Pain clinical practice guidelines \cite{swarm2019adult}, and (3) institutional analgesic dosing protocols from the hospital. Each document was segmented using a fixed-size chunking strategy with a chunk size of 512 tokens and an overlap of 64 tokens. We used the text2vec-base-chinese model as the embedding function $\phi$(·), producing 768-dimensional vectors for both queries and document chunks. The FAISS index was built using an IVF (Inverted File Index) and PQ (Product Quantization) compression for efficient top-k (k=5) retrieval. All retrieved chunks were concatenated with the patient query as contextual input to the LLM.
}
The generator then conditions on both the query and the retrieved documents to produce the output $ y = \text{LLM} (q, d_1, d_2, \dots, d_k).$

The knowledge base $D$ was constructed from curated clinical guidelines, pharmacological references, and institutional pain management protocols. The retrieval module indexed curated medical literature and dosing protocols, enabling the model to access relevant information during inference~\cite{swarm2019adult,caraceni2012use}. This approach was particularly effective in interpreting ambiguous or incomplete medication records and aligning predictions with established clinical standards.

The structure and composition of the input prompt may significantly affect the LLM's ability to generate accurate and clinically meaningful outputs. Therefore, significant effort was dedicated to prompt engineering to develop a standardized, multi-layered query that ensures both replicability and clinical relevance. Three prompt versions were developed, each beginning with the standardized data information and followed by varying levels of detail and guidance. These prompts were designed to elicit predictions of NRS pain scores at 48 and 72 hours, based on physiological indicators, medication records, and clinical context.

Each prompt version was tested on 20 randomly selected patient cases. For each case, the model generated a response that included pain probability classification and explanatory reasoning. Three expert clinicians independently reviewed the responses and annotated them for:
\begin{itemize}
\item \textbf{Interpretability}: clarity, logical flow, and clinical readability
\item \textbf{Completeness}: inclusion of relevant physiological, pharmacological, and contextual factors
\item \textbf{Prediction Accuracy}: correctness of predicted pain risk compared to ground truth NRS scores
\end{itemize}

Prompt variants were evaluated by expert annotators. The final version was selected based on its ability to balance interpretability, completeness, and predicting accuracy. We present representative examples from each prompt version and summarize expert feedback qualitatively, and the final version of the prompt is shown in the Table~\ref{tab:prompt}.
\begin{table}[t]
\caption{Pain Prediction Analysis Prompt}
\label{tab:prompt}
\centering
\setlength{\tabcolsep}{3pt} 
\begin{tabular}{@{}>{\bfseries}l p{0.7\columnwidth}@{}}
\toprule
Component & Requirements \\
\midrule
Patient data & 
\begin{minipage}[t]{0.7\columnwidth}
\{EHR\}, \{Medication Records\}, \{Chief Complaint\}, \{Clinical Notes\}
\end{minipage} \\

Task & 
As a medical expert, please comprehensively analyze the [Patient data], predict the probability of the patient having an NRS pain score $\geq$ 4 in 48h and 72h maintain current medication. Output the results according to the framework: \\
Data Prep. & 
\begin{minipage}[t]{0.7\columnwidth}

1. Key lab/hematologic/metabolic/tumor markers \\
2. Medication analysis: \\
\hspace{0.5em}$\bullet$ Last 24h drugs (name/dose): \_\_\_\_ \\
\hspace{0.5em}$\bullet$ Pharmacodynamics: metabolic risk + inflammation
\end{minipage} \\

Output format& 
\begin{minipage}[t]{0.6\columnwidth}
3. Probability tiers: High ($>$70\%), Medium (30-70\%), Low ($<$30\%)\\
4. Main risk factors: \_\_\_\_
\end{minipage} \\
\bottomrule
\end{tabular}
\end{table}

It begins by establishing a clear data schema, providing the LLM with a standardized set of physiological and biochemical features for each patient case. This is followed by anchoring the model with the patient's baseline clinical state, such as the current NRS pain score, which reframes the task from static prediction to a more clinically relevant trajectory forecast. Crucially, the prompt embeds explicit clinical constraints (i.e., "maintain current medication" ) to reduce ambiguity and ensure the prediction is made under a stable, controlled clinical assumption. Finally, the prompt mandates a structured output format, requiring the LLM to deliver not only a numerical prediction but also a multi-faceted clinical rationale. 

\subsubsection{Pipeline Integration}
Decision-level fusion combined the statistical robustness of machine learning with the contextual depth of the LLM.  Let $p_{\mathrm{ML}}\in[0,1]$ denote the classifier probability.  For marginal confidence $\alpha<p_{\mathrm{ML}}<\beta$, the LLM generated a secondary estimate $p_{\mathrm{LLM}}$.  The final probability was
\[
p_{\text{final}} = \frac{\mathbbm{1}_{\{\alpha < p_{\text{ML}} < \beta\}} \cdot p_{\text{LLM}} + p_{ML}}{1 + \mathbbm{1}_{\{\alpha < p_{\text{ML}} < \beta\}}}
\]
where $\mathbbm 1(\cdot)$ is the indicator function.  Thresholds $\alpha$ and $\beta$ were selected from the development set at the break-even point of precision–recall and set to 0.2 and 0.6, respectively. 
{ 
This formulation ensures that in the marginal confidence region, the final prediction is computed as the average of the ML and LLM outputs, thereby moderating both sources of uncertainty while defaulting to the ML prediction alone when confidence is high.
}
The LLM component was guided by a structured prompt that synthesized laboratory indicators, medication records, and clinical notes. This integration strategy allowed the system to leverage the statistical reliability of ML models and the contextual depth of LLM, particularly in ambiguous or data-sparse scenarios.

\subsection{Evaluation Metrics}
In clinical pain management, both sensitivity and specificity are critical evaluation metrics. Sensitivity reflects the model's ability to identify true pain episodes, minimizing missed alerts that may delay analgesic intervention while specificity measures the ability to avoid false alarms, thereby reducing unnecessary treatments and adverse effects. These considerations are especially relevant in the 72-hour window, where pain-positive cases are sparse and the cost of misclassification is high.
We therefore report four standard classification metrics:
{ 
Sensitivity, Specificity, Accuracy, AUC, and 
\begin{IEEEeqnarray}{rCl}
\begin{aligned}
\mathrm{ECE}
&= \sum_{m=1}^{M} \frac{|B_m|}{N}
\Bigg|
\frac{1}{|B_m|} \sum_{i \in B_m} \mathbb{I}(\hat{y}_i = y_i) \\
&\quad -
\frac{1}{|B_m|} \sum_{i \in B_m} \hat{p}_i
\Bigg|
\end{aligned}
\end{IEEEeqnarray}

\noindent where $N$ denotes the total number of samples,
$M$ is the number of confidence bins,
$B_m$ represents the set of samples whose predicted confidence falls into the
$m$-th bin,
$\hat{p}_i$ is the predicted probability for sample $i$,
$\hat{y}_i$ and $y_i$ denote the predicted and ground-truth labels, respectively,
and $\mathbb{I}(\cdot)$ is the indicator function. And
}
AUC (Area Under the ROC Curve) summarizes the model’s ability to discriminate between pain and non-pain episodes across all classification thresholds, offering a robust measure of overall performance, especially in imbalanced datasets.

\section{Results}
\subsection{ML Performance}
The baseline models, which relied solely on static patient features, demonstrated solid baseline performance. For the 48-hour prediction task, the best-performing model achieved an AUC of 0.886. For the 72-hour prediction, the top model reached an AUC of 0.901. These results are illustrated in Figure~\ref{fig:model_comparison} (a) and (b). 

After incorporating the structured medication dosage variables extracted from unstructured text, model performance improved notably for the 48-hour prediction. The AUC increased to 0.958, indicating a substantial gain in predictive accuracy. For the 72-hour prediction, the AUC remained at 0.901, but the standard deviation decreased from 0.084 to 0.078, suggesting improved model stability and consistency across folds. These updated results are shown in Figure~\ref{fig:model_comparison} (c) and (d). The best-performing models for the 48-hour prediction were Extra Trees (AUC 0.958 ± 0.050, sensitivity 0.840, specificity 0.879), CatBoost (0.942), and LightGBM (0.940). For the 72-hour prediction, CatBoost led with an AUC of 0.901 ± 0.078, followed by Gradient Boosting, Lasso, and LightGBM, which showed slightly lower performance and higher variability.

\begin{figure*}[t]
    \centering

    \begin{subfigure}[t]{0.24\textwidth}
        \centering
        \includegraphics[width=\linewidth, trim=0 0 0 20, clip]{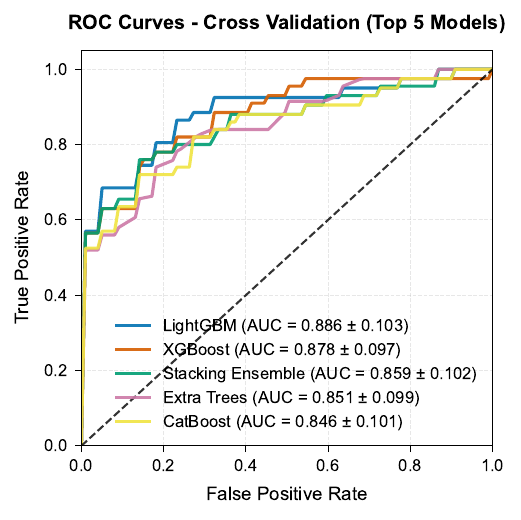}
        \caption{Baseline model - 48h}
        \label{fig:auc_baseline_a}
    \end{subfigure}
    \hfill
    \begin{subfigure}[t]{0.24\textwidth}
        \centering
        \includegraphics[width=\linewidth, trim=0 0 0 20, clip]{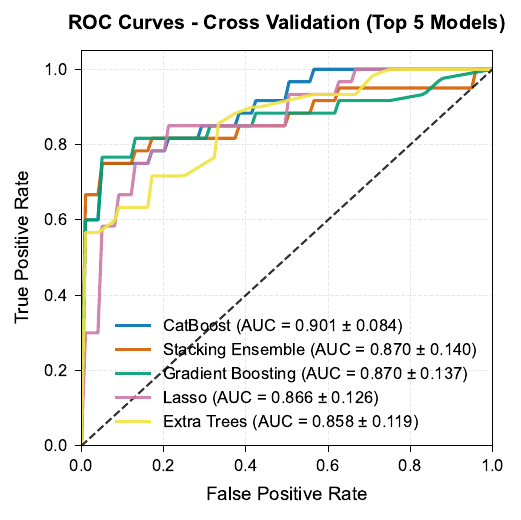}
        \caption{Baseline model - 72h}
        \label{fig:auc_baseline_b}
    \end{subfigure}
    \hfill
    \begin{subfigure}[t]{0.24\textwidth}
        \centering
        \includegraphics[width=\linewidth, trim=0 0 0 20, clip]{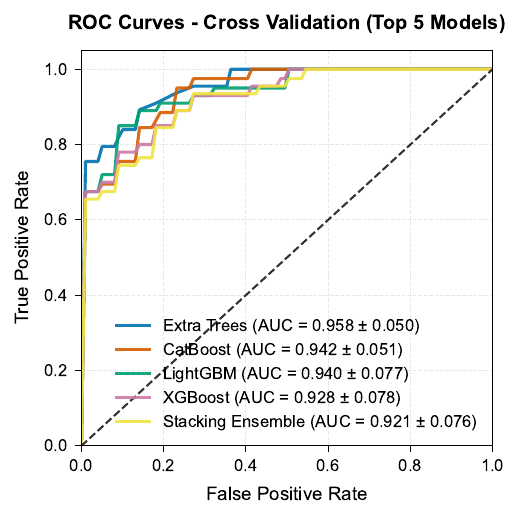}
        \caption{Enhanced model - 48h}
        \label{fig:auc_enhanced_c}
    \end{subfigure}
    \hfill
    \begin{subfigure}[t]{0.24\textwidth}
        \centering
        \includegraphics[width=\linewidth, trim=0 0 0 20, clip]{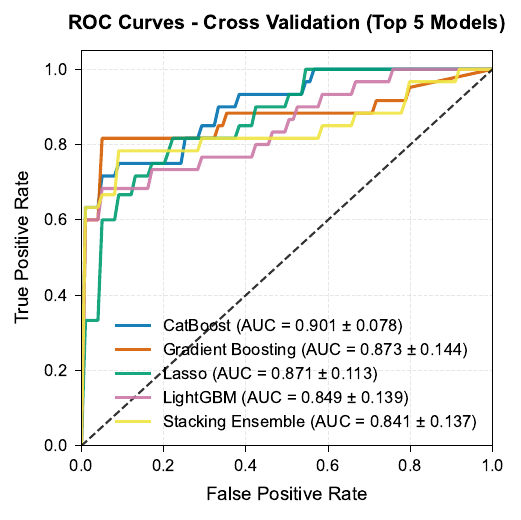}
        \caption{Enhanced model - 72h}
        \label{fig:auc_enhanced_d}
    \end{subfigure}

    \caption{ROC curves from 5-fold cross-validation. Panels (a) and (b) show the baseline model, whereas panels (c) and (d) show the medication-enhanced model.}
    \label{fig:model_comparison}
\end{figure*}

Feature importance was extracted from the final models—Extra Trees for the 48-hour prediction task and CatBoost for the 72-hour task, as shown in Figure~\ref{fig:feature_importance}. For Extra Trees, importance scores were derived from the mean decrease in impurity (MDI), which measures how much each feature contributes to reducing classification error. CatBoost uses a permutation-based approach combined with internal gradient statistics to assess the marginal contribution of each feature to the model’s predictive performance.

\begin{figure*}[t] 
    \centering
    \includegraphics[width=0.9\textwidth]{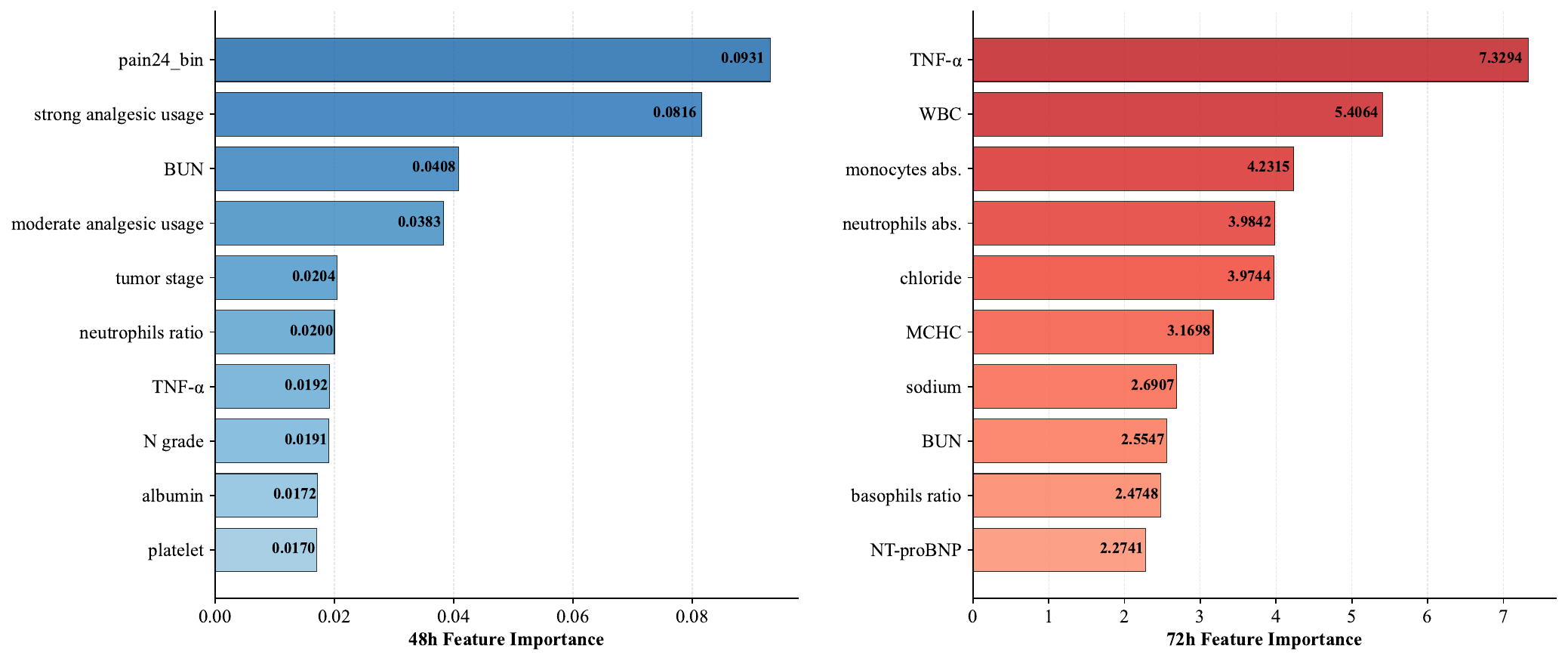}
    \caption{Feature Importance from Extra Trees (48-hour prediction) and CatBoost (72-hour prediction). \textbf{Note.} The numerical values reflect within-model ranking only, not a common unit of effect size across models. BUN = Blood Urea Nitrogen. WBC = White Blood Cell. MCHC = Mean Corpuscular Hemoglobin Concentration. NT-proBNP = N-terminal pro-B-type Natriuretic Peptide. CRP = C-Reactive Protein.}
    \label{fig:feature_importance}
\end{figure*}

The 48-hour model revealed that recent pain status and analgesic exposure were the most influential predictors. The binary indicator of pain within the previous 24 hours (pain24\_bin) had the highest importance score of 0.0931, indicating that recent pain episodes strongly influence short-term risk. The presence of strong opioid usage, derived from structured medication records, followed closely with an importance score of 0.0816. Additional features such as blood urea nitrogen (BUN), moderate opioid usage, and TNF-$\alpha$ contributed moderately, suggesting that metabolic and inflammatory markers also play a role in near-term pain dynamics.

In contrast, the 72-hour model showed a shift toward systemic and inflammatory indicators. TNF-$\alpha$ emerged as the most dominant feature with an importance score of 7.3294, highlighting its role in chronic inflammation and sustained pain. White blood cell count (WBC), monocyte absolute count, and neutrophil absolute count were also highly ranked, with importance scores of 5.406, 4.232, and 3.984 respectively. These hematological markers reflect the body’s immune response and were strongly associated with longer-term pain risk. Electrolyte and metabolic indicators such as chloride, sodium, and albumin showed moderate importance, while tumor-related features including TNM staging and N classification contributed to the model’s understanding of cancer-related pain mechanisms.

Overall, the analysis demonstrates a temporal shift in predictive features. The 48-hour model relies more heavily on immediate clinical observations and medication history, whereas the 72-hour model draws from systemic physiological and inflammatory data. This distinction supports the clinical interpretation that short-term pain is more responsive to recent interventions, while longer-term pain reflects underlying patient conditions.

\subsection{LLM Performance}
The integration of RAG led to improvement in predictive performance of the LLM. When evaluated against the full dataset using ground truth pain labels at 48 and 72 hours, the accuracy increased from 0.650 and 0.662 to 0.744 and 0.774, respectively.
These improvements were most pronounced in cases with sparse, noisy, or incomplete input data, where the retrieved external knowledge helped resolve inconsistencies and provided clinical context that was otherwise missing from structured features alone.

To enhance the model’s predictive accuracy and clinical interpretability, multiple prompt versions were iteratively designed and evaluated by clinical experts. The refinement process focused on improving the model’s ability to reason over complex patient data and generate reliable pain forecasts at 48 and 72 hours.

In the initial trial, a general prompt was used that simply instructed the model to predict pain probability based on patient data. The model responded with fragmented reasoning, referencing physiological indicators such as elevated creatinine (7.23 mg/dL) and TNF-$\alpha$ (68.6 pg/mL), which suggested inflammation and altered drug metabolism. However, the output lacked structure and introduced hallucinated elements, including a fever $\geq$ 38.5°C that was not present in the source data. Experts noted that while the model attempted to interpret lab values, it failed to connect them meaningfully to pain outcomes, resulting in low clinical reliability.

In the second iteration, the prompt was revised to include medication timing and 24-hour NRS scores. This version improved contextual relevance by prompting the model to consider short-acting analgesic effects and pain rebound patterns. For example, the model noted that “short-acting medication effect disappeared” and “first pain score 4,” suggesting a possible increase in pain due to lack of new medication. However, the reasoning remained narrow, focusing almost exclusively on medication timing while ignoring systemic indicators such as CRP, albumin, or metabolic status. Predictions were often inaccurate and lacked explanatory depth, leading experts to rate this version as overly simplistic and inconsistent across cases.

The final version, presented in Table~\ref{tab:prompt}, introduced a structured reasoning framework. The prompt explicitly instructed the model to analyze physiological indicators, inflammatory markers, medication records, and clinical characteristics. This guided the model to integrate multiple dimensions of patient data, resulting in outputs that were well-organized and clinically aligned. For instance, the model identified elevated CRP (35.65 mg/L), normal albumin (38.0 g/L), and poor analgesic response to morphine, concluding a high probability (70\%) of NRS $\geq$ 4 at both time points. Experts consistently rated this version highest for interpretability, completeness, and alignment with ground truth, noting minimal hallucinations and strong clinical relevance.

\subsection{Pipeline Integration}
The combined ML+LLM framework outperformed both individual approaches across key evaluation metrics. For the 48-hour prediction task, the integrated model achieved an accuracy of 0.876, sensitivity of 0.936, and specificity of 0.863. For the 72-hour prediction, accuracy reached 0.917, with sensitivity and specificity at 0.821 and 0.928 respectively. These results indicate that the hybrid approach provides a more balanced and reliable prediction framework, particularly in complex or ambiguous clinical scenarios. This improvement stems from the complementary strengths of ML and LLM—ML captures structured temporal patterns, while LLM interpret nuanced clinical notes—leading to enhanced performance across both short- and long-horizon predictions.

\begin{table}[ht]
\centering
\caption{Ablation Study: Prediction Performance at 48h and 72h}
\label{tab:performance}
{ 
\begin{tabular}{llcccc}
\toprule
\textbf{Time} & \textbf{Metric} 
& \textbf{ML} 
& \textbf{LLM} 
& \textbf{LLM with RAG} 
& \textbf{Hybrid} \\
\midrule
\textbf{48h}
& Sensitivity & 0.830 & 0.723 & 0.851 & 0.936 \\
& Specificity & 0.877 & 0.634 & 0.721 & 0.863 \\
& Accuracy    & 0.868 & 0.650 & 0.744 & \textbf{0.876} \\
& ECE  & 0.132  & 0.265  & 0.213  & \textbf{0.100} \\
\midrule
\textbf{72h}
& Sensitivity & 0.714 & 0.643 & 0.786 & 0.821 \\
& Specificity & 0.933 & 0.664 & 0.773 & 0.928 \\
& Accuracy    & 0.909 & 0.662 & 0.774 & \textbf{0.917} \\
& ECE  & 0.106  & 0.257  & 0.193  & \textbf{0.072} \\
\midrule
\multicolumn{2}{l}{Inference Time (s/case)}
& 0.05 & 0.73 & 0.92 & 0.41 \\
\bottomrule
\end{tabular}
}
\end{table}

To assess whether the observed model performance significantly exceeded chance, we conducted permutation testing using 1,000 iterations of label shuffling for the hybrid model (ML + LLM). For 48h, the observed AUC of 0.958 was significantly higher than the permuted null distribution (mean permuted AUC = 0.500 ± 0.066, p $\textless$ 0.001). Similarly, for 72h, the observed AUC of 0.901 exceeded the null distribution (mean permuted AUC = 0.502 ± 0.078, p $\textless$ 0.001). These results confirm that the learned feature-label associations are not attributable to chance or overfitting artifacts.

\subsection{Performance on the Temporal Validation Cohort}
To assess the generalizability of the hybrid models beyond the original development sample, we applied the model to an independent temporal validation cohort comprising 130 patients recruited during a subsequent enrollment period. The confusion matrices and reliability diagrams for both prediction horizons are presented in Figure~\ref{fig:cm_temporal} and Figure~\ref{fig:reliability_temporal}, respectively.

\begin{figure}[t] 
    \centering
    \includegraphics[width=0.45\textwidth]{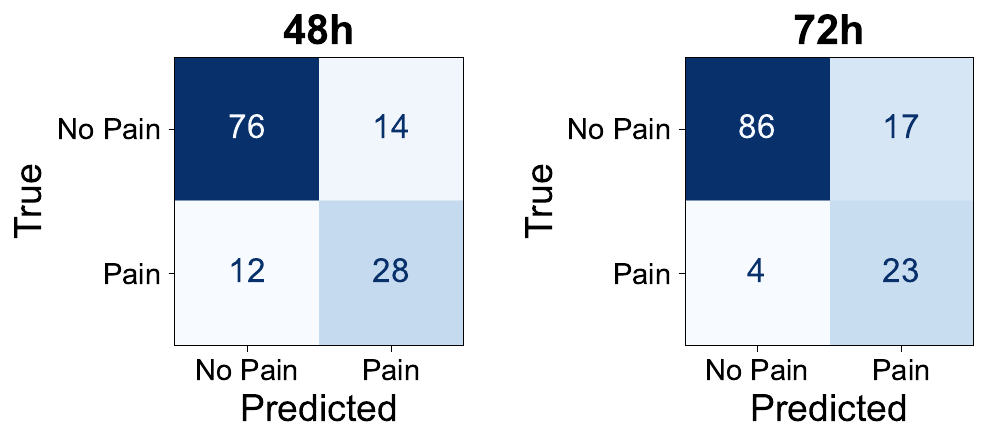}
    \caption{Confusion Matrices for Hybrid Model on Independent Temporal Validation Cohort.}
    \label{fig:cm_temporal}
\end{figure}

\begin{figure*}[t] 
    \centering
    \includegraphics[width=0.75\textwidth]{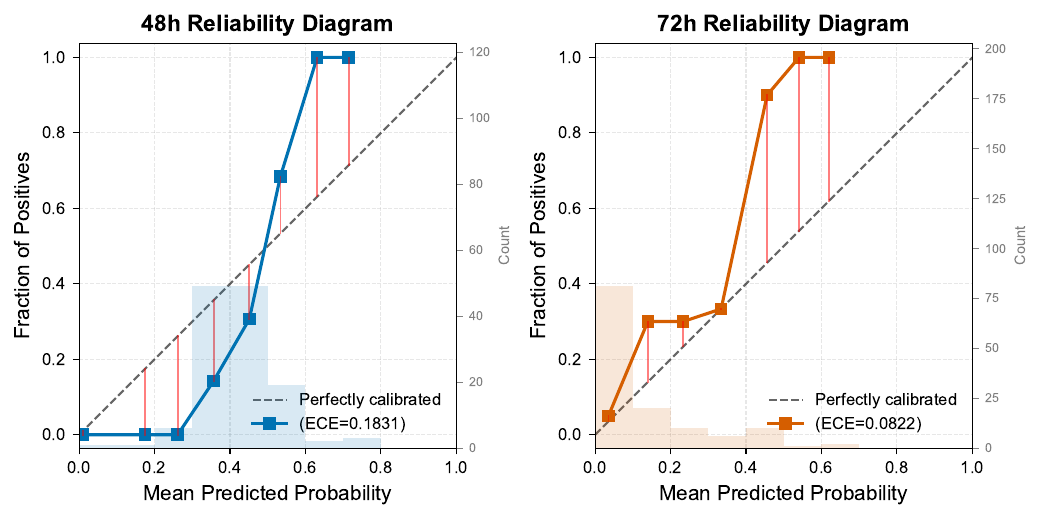}
    \caption{Reliability Diagrams for Hybrid Model on Independent Temporal Validation Cohort.}
    \label{fig:reliability_temporal}
\end{figure*}

Using the previously determined threshold, the 48-hour model achieved an accuracy of 0.800, with a sensitivity of 0.700 and a specificity of 0.844 (AUC = 0.813), as shown in the confusion matrix (Figure~\ref{fig:cm_temporal}, left panel). The reliability diagram (Figure~\ref{fig:reliability_temporal}, left panel) revealed an ECE of 0.183, indicating moderate miscalibration: the model tended to under-predict pain probability in the mid-range bins (0.2–0.4) and exhibited a sharp transition to near-perfect positive fraction above 0.5, suggesting that its predicted probabilities could benefit from post-hoc recalibration (e.g., Platt scaling) in future deployment.

The 72-hour model achieved an accuracy of 0.838, a sensitivity of 0.852, a specificity of 0.835, and an AUC of 0.880 (Figure~\ref{fig:cm_temporal}, right panel). It is worth noting that the model had only 4 false negatives, which is a clinically acceptable margin of error, because the clinical consequences of missing a pain event are more serious than those of a false positive. The reliability diagram (Figure~\ref{fig:reliability_temporal}, right panel) yielded an ECE of 0.082, substantially lower than that of the 48-hour model, indicating that the 72-hour model's predicted probabilities more closely approximated the observed event rates across all probability bins.

Compared with the internal test set results, both models exhibited consistent discriminative performance on the temporal validation cohort. The 72-hour model in particular maintained high sensitivity (0.852) alongside well-calibrated probability estimates, supporting its potential utility as a clinical decision-support tool for identifying patients at elevated risk of sustained pain beyond the initial post-treatment period.

\subsection{Case Studies}
To assess the practical performance of the pain prediction framework, we examined representative cases where predictions from ML, LLM, and their hybrid integration (ML+LLM) diverged as shown in Table~\ref{tab:final_results}.

LLM-only predictions often suffered from low specificity. In several cases, the LLM overpredicted pain based on general risk factors such as metastasis or elevated inflammatory markers, without adequately accounting for actual pain scores or medication response. For instance, in Case 1, the LLM predicted a high probability of pain (0.85) due to chief complaint, bone and brain metastases and elevated IL-6 and TNF-$\alpha$ levels. Similarly, in Cases 3 and 4, the LLM flagged pain based on text in chief complaint, mild inflammation and persistent complaints, overlooking stable analgesic regimens and low NRS scores.

Conversely, ML-only models demonstrated high specificity but missed pain signals embedded in clinical narratives. In Case 2, ML underestimated pain (0.24), failing to capture poor opioid response, which the LLM correctly identified. A similar pattern was seen in Case 7, where ML missed persistent abdominal pain flagged by the LLM.

The integrated ML+LLM model resolved such discrepancies. In Case 6, it preserved ML’s correct prediction despite LLM underestimation. In Case 5, it corrected ML’s false positive using LLM’s contextual understanding. The hybrid approach balanced structured data and narrative interpretation, improving accuracy in complex clinical scenarios.

\begin{table*}[t]
\caption{Cancer Pain Prediction Performance Comparison (NRS $\geq$4)}
\label{tab:final_results}
\centering
\begin{tabular}{>{\centering}m{0.5cm} >{\raggedright}m{5.5cm} >{\raggedright}m{2.5cm} c c c c}
\toprule
\textbf{ID} & \textbf{Clinical Context} & \textbf{Opioid Therapy} & \textbf{True} & \multicolumn{3}{c}{\textbf{Prediction Results}} \\ 
\cmidrule(lr){5-7}
 & & & & \textbf{ML Prob.} & \textbf{LLM Prob.} & \textbf{Integrated} \\ 
\midrule
1 & Pain in right shoulder and back & Oxycodone ER 10mg & No & 0.18 & 0.85 & No \\ 
2 & Found left lung nodules, poorly controlled & Morphine PCA & Yes & 0.24 & 0.95 & Yes \\
3 & Persistent pain on the right side & Oxycodone ER 80mg & No & 0.14 & 0.50 & No \\
4 & Blood in sputum for more than half a month & Oxycodone ER 10mg & No & 0.18 & 0.75 & No \\
5 & ---- & Morphine ER 10mg & No & 0.43 & 0.55 & No \\
6 & ---- & Morphine ER 10mg & Yes & 0.69 & 0.20 & Yes \\
7 & Abdominal cramps & Morphine ER 10mg & Yes & 0.37 & 0.70 & Yes \\
\bottomrule
\end{tabular}
\vspace{0.2cm}

\footnotesize \textbf{Note.} ER: Extended Release; PCA: Patient-Controlled Analgesia\\
\end{table*}

While LLM offer valuable contextual interpretation, they require structured model grounding to avoid overprediction. ML models, on the other hand, benefit from LLM-derived insights in cases where pain indicators are embedded in narrative text. The ML+LLM ensemble consistently demonstrated improved accuracy by balancing both perspectives.

\section{Discussion and Conclusion}

This study introduces a hybrid AI framework that integrates traditional machine learning with a large language model to forecast pain episodes in hospitalized lung cancer patients using real-world EHR data. By combining structured clinical variables with unstructured text interpretation, the model delivers actionable predictions at 48 and 72 hours post-admission (Accuracy = 0.876 and 0.917, respectively), demonstrating strong potential for deployment in real-world hospital workflows.

\textbf{ML Module}.
Incorporating structured representations of medication dosage extracted from clinical text significantly improved short-term pain prediction. The 48-hour models benefited from recent analgesic patterns, especially strong and moderate opioids, indicating that near-term pain risk is sensitive to ongoing treatment. In contrast, medication variables had limited impact on 72-hour predictions, where systemic indicators like inflammatory and hematological markers dominated, reflecting baseline physiological status.

These findings support dynamic model updates, ideally every 24 hours, incorporating real-time analgesic patterns for short-term forecasting and comprehensive physiological profiling for longer-range predictions. The ability to derive predictive features from free-text medication records further underscores the value of structured abstraction from unstructured EHR data.

\textbf{LLM Module}.
The use of LLM enabled flexible and scalable processing of unstructured clinical data. Its ability to integrate diverse information sources without manual preprocessing marked a significant advancement over traditional ML pipelines. The use of RAG significantly enhanced the robustness of the prediction framework.

The modular prompt structure, which segments reasoning into preprocessing, medication analysis, and prediction synthesis, facilitated error tracing, iterative refinement, and transparency, which are particularly valuable in clinical settings. However, standalone LLM predictions exhibited a tendency toward high false positives, particularly in cases with ambiguous complaints or aggressive medication use. This limitation highlighted the need for further refinement or integration with structured models to achieve balanced predictive performance.

\textbf{Decision-Making Value}.
In real-world practice, clinical documentation is often incomplete or ambiguous. Our hybrid approach resolves this by using the ML component to capture structured data patterns and the RAG-augmented LLM to interpret free-text notes. This integration significantly enhances decision support, improving sensitivity in marginal cases and reducing false positives. Notably, the hybrid model reduced missed pain alerts by 10.6\% (48h) and 10.7\% (72h) compared to ML-only baselines, augmenting clinical judgment in data-sparse situations. Furthermore, a key advantage for deployment is inference latency. By restricting LLM invocation to marginal-confidence cases, the hybrid model requires only 0.41s per case, striking an optimal balance between the rapid execution of ML-only models (0.05s) and the computational intensity of full LLM inference (0.92s).

\textbf{Clinical Interpretation of Feature Importance}.
The feature-importance profiles reveal a clinically coherent temporal shift from treatment-related predictors to inflammatory biomarkers as the prediction horizon extends from 48 to 72 hours. In the 48-hour model, prior 24-hour pain status ($\mathrm{pain24\_bin}$, importance $= 0.0931$) and strong analgesic usage ($0.0816$) emerged as the dominant predictors, collectively outweighing the remaining top features combined. This aligns with the well-established observation that recent pain trajectory is the strongest short-term predictor of subsequent pain~\cite{mercadante2015breakthrough}. The prominence of strong and moderate analgesic usage captures both current pain severity and pharmacokinetic context such as dose adequacy and emerging tolerance. BUN (third-ranked) reflects renal clearance of opioid metabolites such as morphine-6-glucuronide~\cite{dean2004opioids}, while tumor stage, N grade, albumin, and platelet count capture disease burden, nutritional reserve, and systemic inflammation, all established correlates of pain intensity~\cite{laird2013prognostic}. Notably, TNF-$\alpha$ and neutrophil ratio already appear at 48h with relatively low importance (0.0192 and 0.0200), foreshadowing the inflammatory dominance at 72h. 
Conversely, the 72-hour model demonstrated a distinct shift toward inflammatory biology. The top four features: TNF-$\alpha$ (7.3294), WBC (5.4064), monocyte absolute count (4.2315), and neutrophil absolute count (3.9842), are all inflammatory markers, while analgesic variables no longer appear in the top ten. This suggests that as the prediction horizon extends, the underlying inflammatory milieu supplants current treatment status as the primary determinant of pain outcomes.

Mechanistically, TNF-$\alpha$ amplifies downstream cytokine cascades, enhances TNFR-dependent nociceptor excitability, and promotes central sensitization~\cite{leung2010tnf, kaye2024tumor}. It is significantly elevated in NSCLC patients with high pain scores~\cite{liu2021expression} and participates in tumor--bone--nerve crosstalk in bone cancer pain~\cite{li2015epigallocatechin}. Moreover, sustained TNF-$\alpha$ upregulation has been implicated in opioid analgesic tolerance~\cite{wang2022clinical, esmaeili2015satureja}, potentially explaining recurrent breakthrough pain within 72 hours of standard dosing.
Neutrophils exacerbate pain via proteases, reactive oxygen species, and pro-nociceptive chemokines~\cite{parisien2022acute}, while monocytes contribute through CCL2/CCR2-mediated migration and subsequent TNF-$\alpha$/IL-1$\beta$ release~\cite{oggero2022dorsal}. The remaining features (chloride, sodium, MCHC, NT-proBNP) likely reflect electrolyte disturbances and physiological stress that indirectly modulate pain perception.
While optimizing analgesic regimens remains critical for short-term pain control, monitoring inflammatory biomarkers, particularly TNF-$\alpha$ and leukocyte subsets, may provide additional prognostic value for identifying patients at risk of sustained or escalating pain over a 72-hour horizon, reflecting the broader role of tumor–immune–neural interactions in cancer pain pathophysiology.

\textbf{Comparison with Existing Literature}.
Contextualizing these metrics against recent systematic reviews of AI in cancer pain management~\cite{salama2024artificial}, our predictive performance (AUC = 0.958 at 48h; AUC = 0.901 at 72h.) compares favorably with recent benchmarks, which report median AUCs of 0.86 and 0.71 for pain prediction and management, respectively. Furthermore, broader literature assessments~\cite{taha2025artificial, ghane2025pain} consistently identify a critical over-reliance on structured data, noting that existing predictive models remain largely siloed from unstructured clinical narratives. Our hybrid ML–LLM architecture directly resolves this modality bottleneck by systematically fusing free-text medication logs and clinical notes with structured electronic health records. However, despite these comparative advantages, our framework shares a ubiquitous limitation identified across the field~\cite{salama2024artificial, taha2025artificial}: the absence of multi-center external validation. Given that only 14\% of analogous studies currently employ external validation cohorts (with overall TRIPOD compliance at 70.7\%), future cross-institutional evaluation remains imperative to ensure robust clinical generalizability.
}

\textbf{Limitations and Future Work}.
This study has some limitations. Although the independent temporal validation cohort demonstrates generalizability across enrollment periods, it does not address potential variation in clinical documentation practices, EHR systems, or patient demographics across institutions.
Future directions include external validation across diverse clinical settings, integration into existing clinical decision support systems, and real-time deployment with feedback loops to enhance practical utility.

\section*{References}

\bibliographystyle{ieeetr}
\bibliography{pain_bib}

\end{document}